\documentclass{article}
\usepackage{nips15submit_e,times}
\usepackage{amsmath,graphicx}
\usepackage[colorlinks=True,citecolor=blue,linkcolor=blue]{hyperref}
\usepackage{hyperref}
\usepackage{url}

\def\x{{\mathbf x}}

\title{Batch Normalized Recurrent Neural Networks}

\author{
C\'{e}sar Laurent \thanks{Equal contribution} \\
Universit\'e de Montr\'eal \\
\And
Gabriel Pereyra $^{*}$  \\
University of Southern California \\
\AND
Phil\'{e}mon Brakel \\
Universit\'e de Montr\'eal \\
\And
Ying Zhang \\
Universit\'e de Montr\'eal \\
\And
Yoshua Bengio \thanks{CIFAR Senior Fellow} \\
Universit\'e de Montr\'eal \\
}

\nipsfinalcopy 

\begin{document}
%
\maketitle
\begin{abstract}
Recurrent Neural Networks (RNNs) are powerful models for sequential data that
have the potential to learn long-term dependencies. However, they are computationally
expensive to train and difficult to parallelize. Recent work has shown that
normalizing intermediate representations of neural networks can significantly
improve convergence rates in feedforward neural networks \cite{ioffe2015}. In
particular, batch normalization, which uses mini-batch statistics to standardize
features, was shown to significantly reduce training time. In this paper, we
show that applying batch normalization to the hidden-to-hidden transitions of our
RNNs doesn't help the training procedure. We also show that when applied to the 
input-to-hidden transitions, batch normalization can lead to a faster convergence of the
training criterion but doesn't seem to improve the generalization performance
on both our language modelling and speech recognition tasks.
All in all, applying batch normalization to RNNs turns out to be more
challenging than applying it to feedforward networks, but certain variants of it
can still be beneficial.
\end{abstract}


\section{Introduction}

Recurrent Neural Networks (RNNs) have received renewed interest due to their
recent success in various domains, including speech recognition
\cite{graves2013speech}, machine translation
\cite{sutskever2014sequence,bahdanau2014} and language modelling \cite{mikolov2012thesis}.
The so-called Long Short-Term Memory (LSTM) \cite{hochreiter1997long} type RNN has been particularly
successful. Often, it seems beneficial to train deep architectures in which
multiple RNNs are stacked on top of each other \cite{graves2013speech}.
Unfortunately, the training cost for large datasets
and deep architectures of stacked RNNs can be prohibitively high, often times an
order of magnitude greater than simpler models like $n$-grams \cite{williams2015scaling}. Because
of this, recent work has explored methods for parallelizing RNNs across multiple
graphics cards (GPUs). In \cite{sutskever2014sequence}, an LSTM type RNN was distributed
layer-wise across multiple GPUs and in \cite{hannun2014deepspeech} a bidirectional RNN was distributed across time. However, due to the sequential nature of RNNs, it is difficult to achieve linear speed ups relative to the number of GPUs.

Another way to reduce training times is through a better conditioned
optimization procedure. Standardizing or whitening of input data has long been
known to improve the convergence of gradient-based optimization methods \cite{lecun2012efficient}.
Extending this idea to multi-layered networks suggests that normalizing or
whitening intermediate representations can similarly improve convergence.
However, applying these transforms would be extremely costly.
In \cite{ioffe2015}, batch normalization was used to standardize intermediate
representations by approximating the population statistics using sample-based
approximations obtained from small subsets of the data, often called mini-batches, that
are also used to obtain gradient approximations for stochastic gradient descent,
the most commonly used optimization method for neural network training.
It has also been shown that convergence can be improved even more by whitening
intermediate representations instead of simply standardizing them \cite{desjardins2015natural}.
These methods reduced the training time of Convolutional Neural Networks (CNNs)
by an order of magnitude and additionallly provided a regularization effect,
leading to state-of-the-art results in object recognition on the ImageNet
dataset \cite{ILSVRC15}. In this paper, we explore how to leverage normalization in RNNs and show that training time can be reduced.


\section{Batch Normalization}

In optimization, feature standardization or whitening is a common procedure
that has been shown to reduce convergence rates \cite{lecun2012efficient}. Extending the idea to deep neural
networks, one can think of an arbitrary layer as receiving samples from
a distribution that is shaped by the layer below. This distribution
changes during the course of training, making any layer but the first
responsible not only for learning a good representation but also for adapting to
a changing input distribution. This distribution variation is termed \textit{Internal
Covariate Shift}, and reducing it is hypothesized to help the training procedure
\cite{ioffe2015}. 

To reduce this internal covariate shift, we could whiten each layer of the network. However,
this often turns out to be too computationally demanding.
Batch normalization \cite{ioffe2015} approximates the whitening by standardizing the intermediate representations using the statistics of the current mini-batch. Given a mini-batch $\mathbf x$, we can calculate the sample mean and sample variance of each feature $k$ along the mini-batch axis

\begin{align}
    \bar \x_k &= \frac{1}{m} \sum_{i=1}^m \x_{i,k},\\
    \sigma^{2}_k &= \frac{1}{m} \sum_{i=1}^m (\x_{i,k} - \bar \x_k)^2,
\end{align}
where $m$ is the size of the mini-batch. Using these statistics, we can standardize each feature as follows

\begin{equation}
	\hat \x_k = \frac{\x_k - \bar \x_k}{\sqrt{\sigma^{2}_{k} + \epsilon}},
\end{equation}
where $\epsilon$ is a small positive constant to improve numerical stability.

However, standardizing the intermediate activations reduces the
representational power of the layer. To account for this, batch normalization introduces additional learnable parameters $\gamma$ and $\beta$, which respectively scale and shift the data, leading to a layer of the form

\begin{equation}
	BN(\mathbf x_k) = \gamma_k \hat \x_k + \beta_k.
\end{equation}
By setting $\gamma_k$ to $\sigma_k$  and $\beta_k$ to $\bar x_k$, the network can recover the original layer representation. So, for a standard feedforward layer in a neural network
\begin{equation}
	\mathbf y = \phi(\mathbf W \mathbf x + \mathbf b),
\end{equation}
where $\mathbf W$ is the weights matrix, $\mathbf b$ is the bias vector, $\mathbf x$ is the input of the layer and $\phi$ is an arbitrary activation function, batch normalization is applied as follows
\begin{equation}
	\mathbf y = \phi(BN(\mathbf W \mathbf x)).
    \label{BNFF}
\end{equation}
Note that the bias vector has been removed, since its effect is cancelled by the standardization.
Since the normalization is now part of the network, the back propagation
procedure needs to be adapted to propagate gradients through the mean and
variance computations as well.

At test time, we can't use the statistics of the mini-batch. Instead, we can estimate them by either forwarding several training mini-batches through the network and averaging their statistics, or by maintaining a running average calculated over each mini-batch seen during training.


\section{Recurrent Neural Networks}

Recurrent Neural Networks (RNNs) extend Neural Networks to sequential data. Given an input sequence of vectors $(\mathbf x_1,\dots, \mathbf x_T)$, they produce a sequence of hidden states $(\mathbf h_1,\dots, \mathbf h_T)$, which are computed at time step $t$ as follows
\begin{equation}
	\mathbf h_t = \phi(\mathbf W_h \mathbf h_{t-1} + \mathbf W_x \x_t),
    \label{RNN}
\end{equation}
where $\mathbf W_h$ is the recurrent weight matrix, $\mathbf W_x$ is the input-to-hidden weight matrix, and $\phi$ is an arbitrary activation function. 

If we have access to the whole input sequence, we can use information not only
from the past time steps, but also from the future ones, allowing for
bidirectional RNNs \cite{schuster1997bidirectional} 
\begin{align}
	\overrightarrow{\mathbf h}_t &= \phi(\overrightarrow{\mathbf W}_h \overrightarrow{\mathbf h}_{t-1} + \overrightarrow{\mathbf W}_x \x_t),\\
	\overleftarrow{\mathbf h}_t &= \phi(\overleftarrow{\mathbf W}_h \overleftarrow{\mathbf h}_{t+1} + \overleftarrow{\mathbf W}_x \x_t),\\
	\mathbf h_t &= [\overrightarrow{\mathbf h}_t : \overleftarrow{\mathbf h}_t],
\end{align}
where $[\mathbf x : \mathbf y]$ denotes the concatenation of $\mathbf x$ and
$\mathbf y$. Finally, we can stack RNNs by using $\mathbf h$ as the input to
another RNN, creating deeper architectures \cite{pascanu2013construct}
\begin{equation}
	\mathbf h_t^{l} = \phi(\mathbf W_h \mathbf h_{t-1}^{l} + \mathbf W_x \mathbf h_t^{l-1}).
\end{equation}

In vanilla RNNs, the activation function $\phi$ is usually a sigmoid
function, such as the hyperbolic tangent. Training such networks is known to be
particularly difficult, because of vanishing and exploding gradients \cite{pascanu2012difficulty}.

\subsection{Long Short-Term Memory}

A commonly used recurrent structure is the Long Short-Term Memory (LSTM). It
addresses the vanishing gradient problem commonly found in vanilla RNNs by
incorporating gating functions into its state dynamics \cite{hochreiter1997long}. At each time step, an LSTM maintains a hidden vector $\mathbf h$ and a cell vector $\mathbf c$  responsible for controlling state updates and outputs. More concretely, we define the computation at time step $t$
as follows \cite{gers2003learning}:
\begin{align}
    \mathbf i_t &= \text{sigmoid}(\mathbf W_{hi} \mathbf h_{t-1} + \mathbf W_{xi} \x_t)\\
	\mathbf f_t &= \text{sigmoid}(\mathbf W_{hf} \mathbf h_{t-1} + \mathbf W_{hf} \x_t)\\
	\mathbf c_t &= \mathbf f_t \odot \mathbf c_{t-1} + \mathbf i_t  \odot \tanh(\mathbf W_{hc} \mathbf h_{t-1} + \mathbf W_{xc} \x_t)\\
	\mathbf o_t &= \text{sigmoid}(\mathbf W_{ho} \mathbf h_{t-1} + \mathbf W_{hx} \x_t + \mathbf W_{co} \mathbf c_t)\\
	\mathbf h_t &= \mathbf o_t \odot \tanh(\mathbf c_t) 
\end{align}
where $\text{sigmoid}(\cdot)$ is the logistic sigmoid function, $\tanh$ is the
hyperbolic tangent function, $\mathbf W_{h\cdot}$ are the recurrent weight
matrices and $\mathbf W_{x\cdot}$ are the input-to-hiddent weight matrices. $\mathbf i_t$, $\mathbf f_t$ and $\mathbf o_t$ are respectively the input, forget and output gates, and $\mathbf c_t$ is the cell.


\section{Batch Normalization for RNNs}

From equation~\ref{BNFF}, an analogous way to apply batch normalization to an
RNN would be as follows:
\begin{equation}
	\mathbf h_t = \phi(BN(\mathbf W_h \mathbf h_{t-1} + \mathbf W_x \x_t)).
    \label{BNRNN}
\end{equation}
However, in our experiments, when batch normalization was applied in this
fashion, it didn't help the training procedure (see appendix~\ref{appendix} for more details). Instead we propose to apply batch normalization only to the input-to-hidden transition ($\mathbf W_x \mathbf x_t$), i.e. as follows:
\begin{equation}
	\mathbf h_t = \phi(\mathbf W_h \mathbf h_{t-1} + BN(\mathbf W_x \x_t)).
\end{equation}
This idea is similar to the way dropout \cite{srivastava2014dropout} can be
applied to RNNs \cite{zaremba2014recurrent}: batch normalization is applied only
on the vertical connections (i.e. from one layer to another) and not on the
horizontal connections (i.e. within the recurrent layer). We use the same
principle for LSTMs: batch normalization is only applied after multiplication
with the input-to-hidden
weight matrices $\mathbf W_{x\cdot}$.

\subsection{Frame-wise and Sequence-wise Normalization}

In experiments where we don't have access to the future frames, like in language modelling where the goal is to predict the next character, we are forced to compute the normalization a each time step
\begin{equation}
	\hat \x_{k,t} = \frac{\x_{k,t} - \bar \x_{k,t}}{\sqrt{\sigma^{2}_{k,t} + \epsilon}}.
\end{equation}
We'll refer to this as \emph{frame-wise normalization}.

In applications like speech recognition, we usually have access to the entire sequences. However, those sequences may have variable length. Usually, when using mini-batches, the smaller sequences are padded with zeroes to match the size of the longest sequence of the mini-batch. In such setups we can't use frame-wise normalization, because the number of unpadded frames decreases along the time axis, leading to increasingly poorer statistics estimates. To solve this problem, we apply a sequence-wise normalization, where we compute the mean and variance of each feature along both the time and batch axis using
\begin{align}
    \bar \x_k &= \frac{1}{n} \sum_{i=1}^m \sum_{t=1}^T \x_{i,t,k},\\
    \sigma^{2}_k &= \frac{1}{n} \sum_{i=1}^m \sum_{t=1}^T (\x_{i,t,k} - \bar \x_k)^2,
\end{align}
where $T$ is the length of each sequence and $n$ is the total number of unpadded frames in the mini-batch.
We'll refer to this type of normalization as \emph{sequence-wise normalization}.


\section{Experiments}

We ran experiments on a speech recognition task and a language modelling task.
The models were implemented using Theano \cite{Bastien-Theano-2012} and Blocks \cite{blocksfuel}.

\subsection{Speech Alignment Prediction}

For the speech task, we used the Wall Street Journal (WSJ) \cite{paul1992design}
speech corpus. We used the si284 split as training set and evaluated our models
on the dev93 development set. The raw audio was transformed into 40 dimensional
log mel filter-banks (plus energy), with deltas and delta-deltas. As in
\cite{graves2013hybrid}, the forced alignments were generated from the Kaldi
recipe tri4b, leading to 3546 clustered triphone states. Because of memory issues, we removed from the training set the sequences that were longer than 1300 frames (4.6\% of the set), leading to a training set of 35746 sequences.

The baseline model (BL) is a stack of 5 bidirectional LSTM layers with 250
hidden units each, followed by a size 3546 softmax output layer. All the weights
were initialized using the Glorot \cite{glorot2010understanding} scheme and all the biases were set to zero. For the batch normalized model (BN) we applied sequence-wise normalization to each LSTM of the baseline model.
Both networks were trained using standard SGD with momentum, with a fixed learning rate of 1e-4 and a fixed momentum factor of 0.9. The mini-batch size is 24.
 
\begin{figure}[t]
  \centering
  \includegraphics[width=0.75\textwidth]{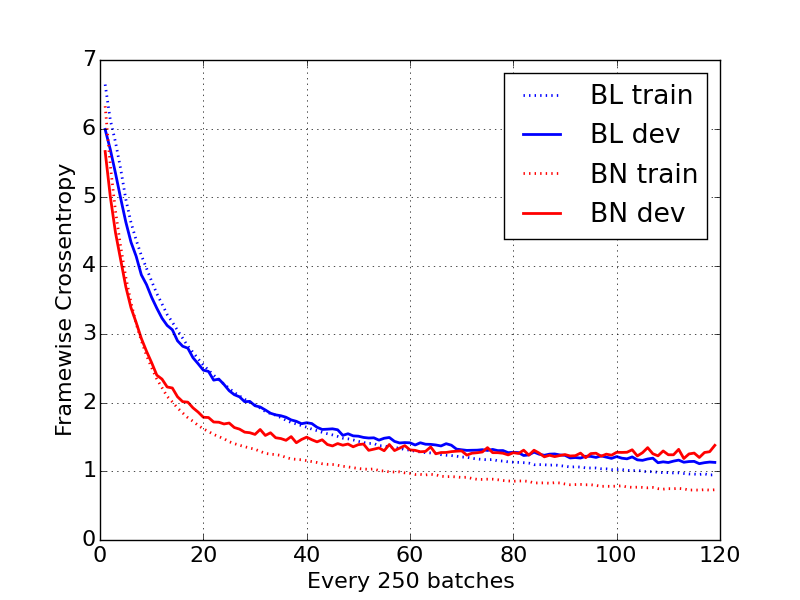}
  \caption{Frame-wise cross entropy on WSJ for the baseline (blue) and batch normalized (red) networks. The dotted lines are the training curves and the solid lines are the validation curves.}
  \label{speech_experiment}
\end{figure}

\begin{table}
	\begin{center}
		\begin{tabular}{lcccc}
  			\hline
	
			\bf Model & \multicolumn{2}{c}{\bf Train} & \multicolumn{2}{c}{\bf Dev} \\
			
			\hline
			
           		 & FCE & FER & FCE & FER \\
            
			\hline
						
  			BiRNN    & 0.95 & 0.28 & \bf 1.11 & \bf 0.33 \\
  			BiRNN (BN) & \bf 0.73 & \bf 0.22 & 1.19 & 0.34 \\
  			
			\hline
		\end{tabular}
		\caption{Best framewise cross entropy (FCE) and frame error rate (FER) on the training and development sets for both networks.}
        \label{speech_table}
	\end{center}
\end{table}

\subsection{Language Modeling}

We used the Penn Treebank (PTB) \cite{marcus1993building} corpus for our
language modeling experiments. We use the standard split (929k training words,
73k validation words, and 82k test words) and vocabulary of 10k words. We train
a small, medium and large LSTM as described in \cite{zaremba2014recurrent}.

All models consist of two stacked LSTM layers and are trained with stochastic gradient descent (SGD) with a learning rate of 1 and a mini-batch size of 32.

The small LSTM has two layers of 200 memory cells, with parameters being initialized from a uniform distribution with range  [-0.1, 0.1]. We back propagate across 20 time steps and the gradients are scaled according to the maximum norm of the gradients whenever the norm is greater than 10. We train for 15 epochs and halve the learning rate every epoch after the 6th.  

The medium LSTM has a hidden size of 650 for both layers, with parameters being initialized from a uniform distribution with range [-0.05, 0.05]. We apply dropout with probability of 50\% between all layers. We back propagate across 35 time steps and gradients are scaled according to the maximum norm of the gradients whenever the norm is greater than 5. We train for 40 epochs and divide the learning rate by 1.2 every epoch after the 6th.

The Large LSTM has two layers of 1500 memory cells, with parameters being initialized from a uniform distribution with range [-0.04, 0.04]. We apply dropout between all layers. We back propagate across 35 time steps and gradients are scaled according to the maximum norm of the gradients whenever the norm is greater than 5. We train for 55 epochs and divide the learning rate by 1.15 every epoch after the 15th.

\begin{figure}[t]
  \centering
  \includegraphics[width=0.75\textwidth]{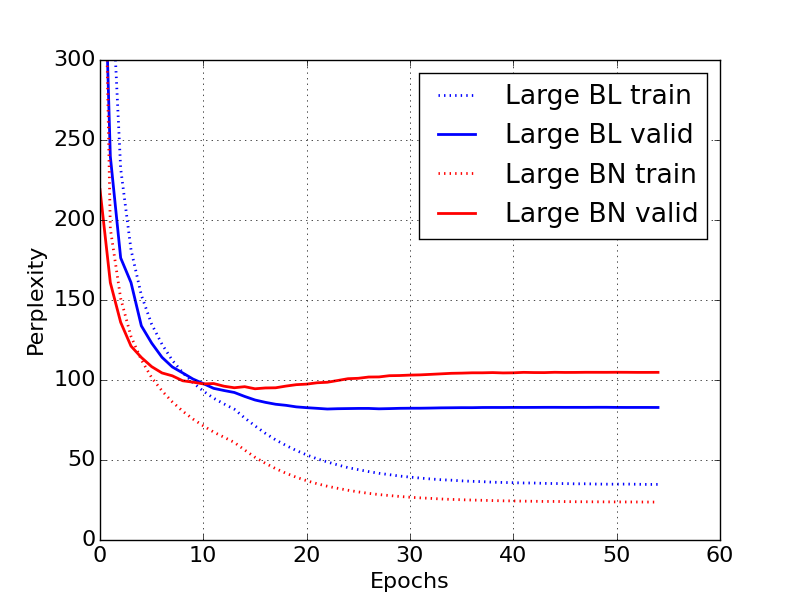}
  \caption{Large LSTM on Penn Treebank for the baseline (blue) and the batch normalized (red) networks. The dotted lines are the training curves and the solid lines are the validation curves.}
  \label{language_experiment}
\end{figure}

{\renewcommand{\arraystretch}{1.4}

\begin{table}
	\begin{center}
		\begin{tabular}{lcr}
  			\hline
  			
			\multicolumn{1}{l}{\bf Model}  &\multicolumn{1}{c}{\bf Train}  &\multicolumn{1}{c}{\bf Valid} \\
  			
			\hline
  			
  			Small LSTM		& 78.5 & 119.2 \\
  			Small LSTM (BN)	& 62.5 & 120.9 \\

  			\hline

  			Medium LSTM		& 49.1 & 89.0 \\
  			Medium LSTM (BN)	& 41.0 & 90.6	  \\

  			\hline

  			Large LSTM		& 49.3 & \bf 81.8 \\
  			Large LSTM (BN)	& \bf 35.0 & 97.4 \\ 

  			\hline

		\end{tabular}
	
		\caption{Best perplexity on training and development sets for LSTMs on Penn Treebank.}
        \label{language_table}
	\end{center}
\end{table}


\section{Results and Discussion}

Figure~\ref{speech_experiment} shows the training and development framewise
cross entropy curves for both networks of the speech experiments. As we can see,
the batch normalized networks trains faster (at some points about twice as fast as the
baseline), but overfits more. The best results, reported in table~\ref{speech_table}, are comparable to the ones obtained in \cite{graves2013hybrid}.

Figure~\ref{language_experiment} shows the training and validation perplexity for the large LSTM network of the language experiment. We can also observe that the training is faster when we apply batch normalization to the network. However, it also overfits more than the baseline version. The best results are reported in table~\ref{language_table}.

For both experiments we observed a faster training and a greater overfitting
when using our version of batch normalization. This last effect is less
prevalent in the speech experiment, perhaps because the training set is way
bigger, or perhaps because the frame-wise normalization is less effective than the sequence-wise one. It can also be caused by the experimental setup: in the language modeling task we predict one character at a time, whereas we predict the whole sequence in the speech experiment.

Batch normalization also allows for higher learning rates in feedforward networks, however since we only applied batch normalization to parts of the network, higher learning rates didn't work well because they affected un-normalized parts as well.


Our experiments suggest that applying batch normalization to the input-to-hidden connections in RNNs can improve the conditioning of the optimization problem. Future directions include whitening input-to-hidden connections \cite{desjardins2015natural} and normalizing the hidden state instead of just a portion of the network.

\subsubsection*{Acknowledgments}
Part of this work was funded by Samsung. We also want to thank Nervana Systems for providing GPUs.

\bibliographystyle{IEEEbib}

\bibliography{paperrefs}


\appendix
\section{Experimentations with Normalization Inside the Recurrence}
\label{appendix}

In our first experiments we investigated if batch normalization can be applied in the same way as in a feedforward network (equation~\ref{BNRNN}). We tried it on a language modelling task on the PennTreebank dataset, where the goal was to predict the next characters of a fixed length sequence of 100 symbols.

The network is composed of a lookup table of dimension 250 followed by 3 layers
of simple recurrent networks with 250 hidden units each. A dimension 50 softmax layer is added on the top. In the batch normalized networks, we apply batch normalization to the hidden-to-hidden transition, as in equation~\ref{BNRNN}, meaning that we compute one mean and one variance for each of the 250 features at each time step. For inference, we also keep track of the statistics for each time step. However, we used the same $\gamma$ and $\beta$ for each time step.

The lookup table is randomly initialized using an isotropic Gaussian with zero mean and unit variance. All the other matrices of the network are initialized using the Glorot scheme \cite{glorot2010understanding} and all the bias are set to zero. We used SGD with momentum. We performed a random search over the learning rate (distributed in the range [0.0001, 1]), the momentum (with possible values of 0.5, 0.8, 0.9, 0.95, 0.995), and the batch size (32, 64 or 128). We let the experiment run for 20 epochs. A total of 52 experiments were performed.

\begin{figure}[t]
  \centering
  \includegraphics[width=0.75\textwidth]{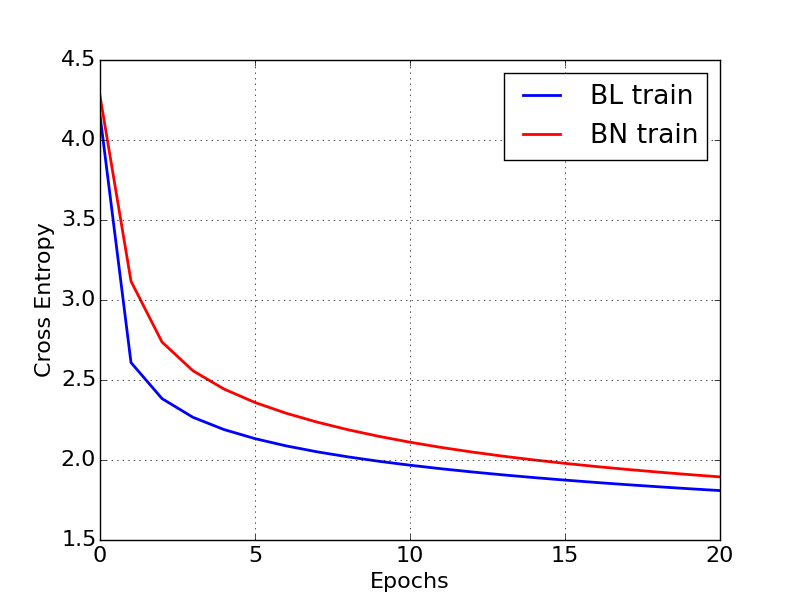}
  \caption{Typical training curves obtained during the grid search. The baseline network is in blue and batch normalized one in red. For this experiment, the hyper-parameters are: learning rate 7.8e-4, momentum 0.5, batch size 64.}
  \label{first_experiment}
\end{figure}

In every experiment that we ran, the performances of batch normalized networks
were always slightly worse than (or at best equivalent to) the baseline
networks, except for the ones where the learning rate is too high and the
baseline diverges while the batch normalized one is still able to train.
Figure~\ref{first_experiment} shows an example of a working experiment. We
observed that in practically all the experiments that converged, the
normalization was actually harming the performance. Table~\ref{first_table} shows the results of the best baseline and batch normalized networks. We can observe that both best networks have similar performances. The settings for the best baseline are: learning rate 0.42, momentum 0.95, batch size 32. The settings for the best batch normalized network are: learning rate 3.71e-4, momentum 0.995, batch size 128. 

Those results suggest that this way of applying batch normalization in the recurrent networks is not optimal. It seems that batch normalization hurts the training procedure. It may be due to the fact that we estimate new statistics at each time step, or because of the repeated application of $\gamma$ and $\beta$ during the recurrent procedure, which could lead to exploding or vanishing gradients. We will investigate more in depth what happens in the batch normalized networks, especially during the back-propagation.

\begin{table}[t]
	\begin{center}
		\begin{tabular}{lcr}
  			\hline
  			
			\multicolumn{1}{l}{\bf Model}  &\multicolumn{1}{c}{\bf Train}  &\multicolumn{1}{c}{\bf Valid} \\
  			
			\hline
  			
  	        Best Baseline   & 1.05 & 1.10 \\
  		    Best Batch Norm & 1.07 & 1.11 \\ 

  			\hline

		\end{tabular}
	
		\caption{Best frame-wise crossentropy for the best baseline network and for the best batch normalized one.}
        \label{first_table}
	\end{center}
\end{table}

\end{document}